%% file: main.tex
\documentclass[10pt,twocolumn,letterpaper]{article}

\usepackage[pagenumbers]{iccv} % To force page numbers, e.g. for an arXiv version

\usepackage{wrapfig}
\usepackage{placeins}
\usepackage{url}
\usepackage{graphicx}
\usepackage{multirow}
\usepackage{algpseudocode}  % Algorithmic notation
\usepackage{adjustbox}
\usepackage{amsmath}
\usepackage{bbm}
\usepackage{float}
\usepackage{rotating} % Rotating 
\PassOptionsToPackage{table,dvipsnames}{xcolor}
\usepackage{colortbl} 

\usepackage{algorithm}
\usepackage{setspace}
\usepackage{booktabs}

\newcommand{\set}[1]{\mathcal{#1}}


\definecolor{iccvblue}{rgb}{0.21,0.49,0.74}
\usepackage[pagebackref,breaklinks,colorlinks,allcolors=iccvblue]{hyperref}

\def\paperID{7488} % *** Enter the Paper ID here
\def\confName{ICCV}
\def\confYear{2025}

\title{Robust Distribution Alignment for Industrial Anomaly Detection under Distribution Shift}

\author{\textbf{Jingyi Liao\textsuperscript{1,2}, Xun Xu\textsuperscript{1}, Yongyi Su\textsuperscript{3}, Rong-Cheng Tu\textsuperscript{2}, Yifan Liu\textsuperscript{4}, Dacheng Tao\textsuperscript{2}, Xulei Yang\textsuperscript{1}}\\ \\
\textsuperscript{1}Institute for Infocomm Research, A*STAR  
\textsuperscript{2}Nanyang Technological University \\
\textsuperscript{3}South China University of Technology  
\textsuperscript{4}National University of Singapore}
\begin{document}
\maketitle
\input{sec/0_abstract}    
\input{sec/1_intro}
\input{sec/2_relatedwork}
\input{sec/3_method}
\input{sec/4_experiments}
\input{sec/5_conclusion}
{
    \small
    \bibliographystyle{ieeenat_fullname}
    \bibliography{main}
}

\end{document}

%% file: sec/0_abstract.tex
\begin{abstract}
Anomaly detection plays a crucial role in quality control for industrial applications. However, ensuring robustness under unseen domain shifts such as lighting variations or sensor drift remains a significant challenge. Existing methods attempt to address domain shifts by training generalizable models but often rely on prior knowledge of target distributions and can hardly generalise to backbones designed for other data modalities. To overcome these limitations, we build upon memory-bank-based anomaly detection methods, optimizing a robust Sinkhorn distance on limited target training data to enhance generalization to unseen target domains. We evaluate the effectiveness on both 2D and 3D anomaly detection benchmarks with simulated distribution shifts. Our proposed method demonstrates superior results compared with state-of-the-art anomaly detection and domain adaptation methods.
% Anomaly detection is essential for quality control in industrial applications. The robustness of industrial anomaly detection under unseen domain shift, such as changes in lighting or sensor drift, remains a critical challenge. Existing works address domain shifts by training a generalizable model. However, reliance on prior knowledge of target distributions, among others, limits the practicality of the existing methods. To overcome these limitations, we build upon memory-bank based anomaly detection methods. We optimize a robust Sinkhorn distance on small target training data to improve the robustness against unseen target testing data.
% To overcome these limitations, we build a robust anomaly detection method based on PatchCore, a memory bank-based anomaly detection method. We improve PatchCore against \textcolor{black}{test-time distribution} shift through a robust distribution alignment which leverages a discretised Sinkhorn distance with target data augmentation. 
% We evaluate the effectiveness on both 2D and 3D anomaly detection benchmarks with simulated distribution shifts. Our proposed method demonstrates superior results compared with state-of-the-art anomaly detection and domain adaptation methods.
\end{abstract}

%% file: sec/1_intro.tex
\section{Introduction}
\label{sec:intro}

\begin{figure}[htbp]
    \centering
    % \captionsetup{justification=centering}
    
    \begin{subfigure}[b]{\linewidth} 
        \centering
        \includegraphics[width=0.95\linewidth]{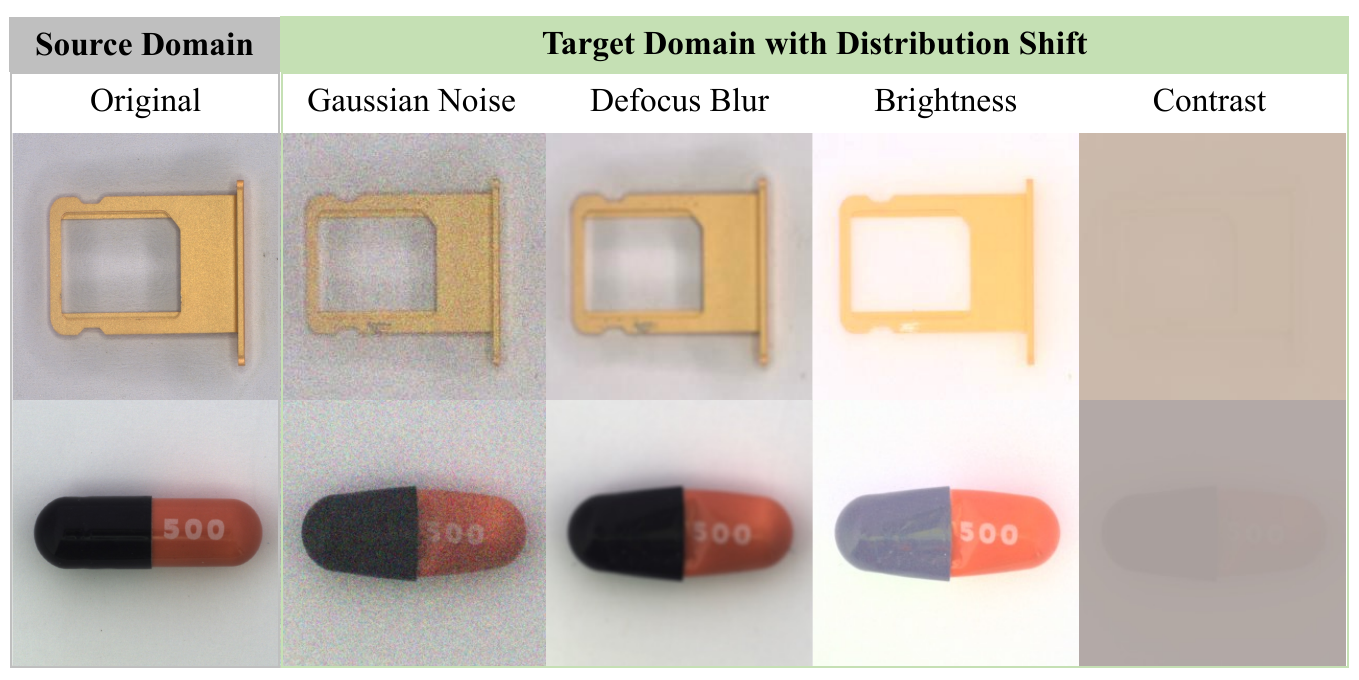} 
        % \caption{Illustrations of the synthesized distribution shift on RealIAD (upper) and MVTec (lower) dataset.}
        \label{fig:samples_small}
    \end{subfigure}
    
    % \vspace{1em}
    \vspace{-0.3cm}
    \begin{subfigure}[b]{\linewidth} 
        \centering
        \includegraphics[width=0.95\linewidth]{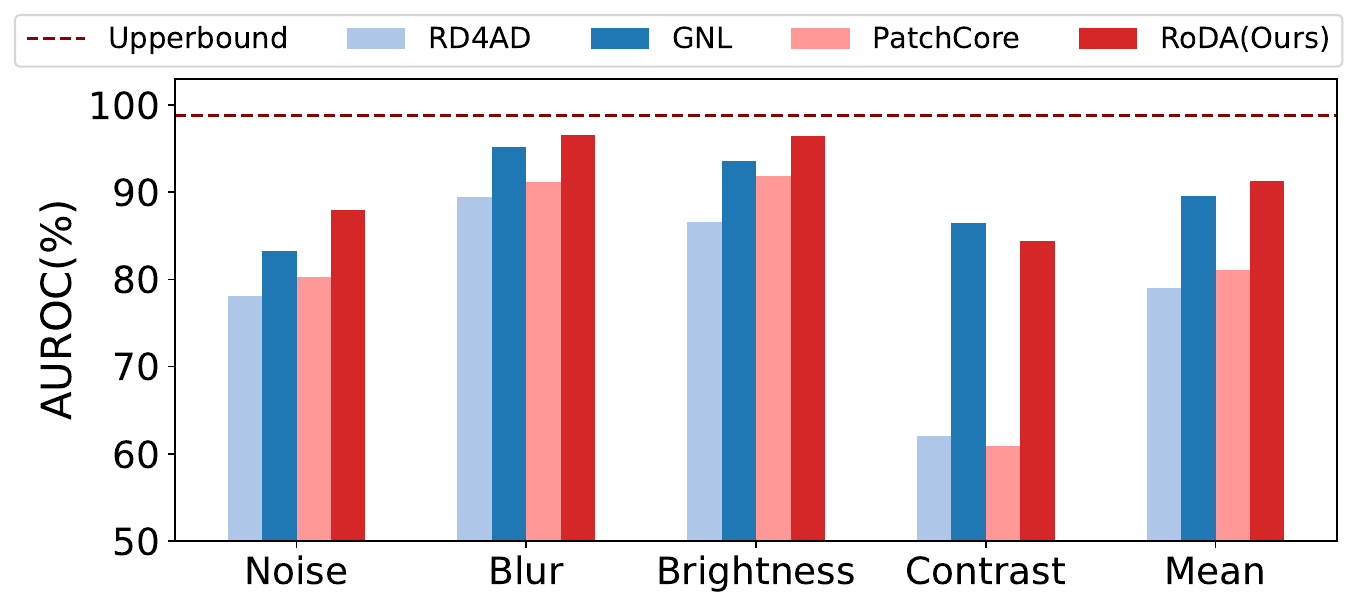} 
        % \caption{Upperbound is }
        \label{fig:barchart}
    \end{subfigure}
    \vspace{-0.6cm}
    \caption{Upper: Examples of synthesized distribution shifts~\cite{hendrycks2019benchmarking} on MVTec and RealIAD. Lower: Results from top baselines and our method. The Upperbound is trained and tested on the original distribution. GNL~\cite{cao2023anomaly} performs well on ``Contrast" shift due to AutoContrast augmentation during training.}
    \vspace{-0.5cm}
    \label{fig:intro_figure}
\end{figure}

%Detecting anomalous patterns is essential for maintaining quality control in industrial applications. Current state-of-the-art methods for industrial anomaly detection typically require extensive defect-free training samples to model the distribution of normal patterns, often using generative models~\cite{deng2022anomaly, zhang2023unsupervised} or memory banks~\cite{roth2022towards, xie2023pushing, gu2023remembering, hu2024dmad}. These approaches have demonstrated impressive performance across various industrial anomaly detection datasets, creating the impression that the challenge has been largely addressed. However, a critical aspect often overlooked is the robustness of these methods, which is essential for real-world applications. In particular, domain shift, where training and testing data distributions differ due to factors like changes in lighting or sensor drift, is a frequent challenge in industrial settings. Examples of synthesized distribution shift~\cite{hendrycks2019benchmarking} are illustrated in Fig.~\ref{fig:intro_figure}. The performances of state-of-the-art methods are compromised by the distribution shifts, evidenced by the significant gap between the ``Upperbound'' (results on clean testing samples) and each method on corrupted testing data.

% similar to source-light/source-relaxed setting.
% only access a few images from target domain, with anomaly samples unfiltered. 

Detecting anomalous patterns is essential for maintaining quality control in industrial applications. Current state-of-the-art methods for industrial anomaly detection typically require extensive defect-free training samples to model the distribution of normal patterns. These methods often employ reconstruction~\cite{deng2022anomaly, zhang2023unsupervised} or memory banks~\cite{roth2022towards, xie2023pushing, gu2023remembering, hu2024dmad} to capture normal data distributions. While demonstrating impressive performance, they often overlook the robustness in real-world applications. In particular, domain shift, where training/source and testing/target data distributions~\footnote{We respectively use training/source and testing/target interchangeably throughout the manuscript.} differ due to factors like changes in lighting or sensor drift, remain a significant challenge in industrial settings. Examples of synthesized distribution shift~\cite{hendrycks2019benchmarking} are illustrated in Fig.~\ref{fig:intro_figure} (upper). The performances of state-of-the-art methods are significantly compromised by the distribution shifts, as evidenced by the gap between the ``Upperbound'' (results on clean testing data) and their performance on corrupted testing data.

% enphsize the importance the setting while only limited works
% To our knowledge, generalized normality learning (GNL)~\cite{cao2023anomaly} is the sole pioneering work that addresses this issue by framing anomaly detection under distribution shift as an out-of-distribution (OOD) generalization problem.
% The most related pioneering work, generalized normality learning (GNL)~\cite{cao2023anomaly}, 

Recent works~\cite{cao2023anomaly, kashiani2024roads} tackle this issue by treating anomaly detection under distribution shift as an out-of-distribution (OOD) generalization problem. 
During training, they encourage consistency in the intermediate features of normal samples under different data augmentations, which is sometimes referred to as synthetic OOD by~\cite{kashiani2024roads}. Training model with this objective ensures that the model's representation is less sensitive to shifts in the data distribution at test time. %For inference, GNL utilizes exact feature distribution matching (EFDM)~\cite{zhang2022exact} to align testing samples with randomly sampled normal data from the training set, achieving superior results on corrupted test datasets.

Despite recent advancements, domain generalization approaches face several key limitations. First, they rely on data augmentations or synthetic OOD samples that mimic target distribution shifts during training. This inherently \textit{assumes prior knowledge of the target domain shift}, which may not always hold. For example, GNL~\cite{cao2023anomaly} suffers performance degradation when the actual distribution shift is not well represented by training augmentations.
Second, these methods are built on reverse distillation~\cite{deng2022anomaly}, which requires \textit{constructing a decoder to invert the encoder}. This design choice limits flexibility, making it non-trivial to extend to architectures such as 3D point cloud encoders.
Finally, GNL~\cite{cao2023anomaly} necessitates \textit{access to a random normal training sample} at inference. However, privacy constraints and storage limitations often restrict access to training data, making this requirement impractical in real-world scenarios.

% First, GNL is built upon reverse distillation~\cite{deng2022anomaly} which requires building a \textit{decoder by reversing the encoder}. It is not trivial to adapt this method with arbitrary backbone networks, e.g. 3D point cloud encoder. Second, GNL requires \textit{access to a random normal training sample} at inference stage. Training samples may be restricted due to privacy concerns or data storage constraints, rendering the method less practical. 
% % as shown in figure 1, the contrast augmentation is introduced in xxx
% Finally, GNL introduced data augmentation resembling target data distribution at training stage. Such a practice requires \textit{knowing the potential domain shift at training stage}. As shown in our evaluations, GNL's performance may degrade when the distribution shift at test time is unknown during the training stage for data augmentation. %Thus, a more flexible approach to industrial anomaly detection is needed. 
% clearence of difference between training sample and features. 

%We propose two critical constraints for an effective solution: i) no retraining or modification of the training process, and ii) no access to training data during inference.

\begin{figure*}
    \centering
    \includegraphics[width=0.9\linewidth]{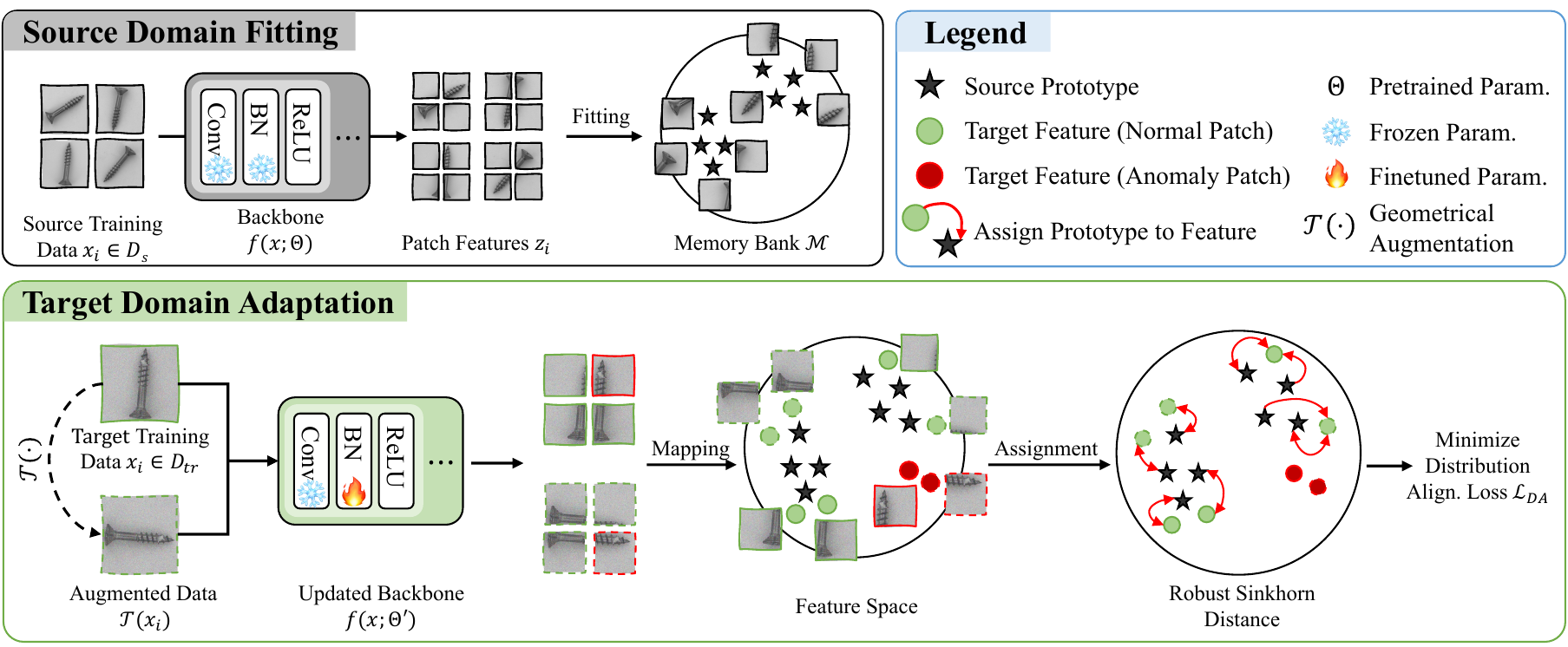}
    \vspace{-0.3cm}
    \caption{Illustration of pipeline of RoDA. The source domain fitting stage constructs a memory bank of normal training features, which serves as a reference for anomaly detection. In the domain adaptation stage, a limited amount of target domain data is augmented and aligned with the source memory bank through robust optimal transport.}
    \vspace{-0.5cm}   
    \label{fig:framework}
\end{figure*}

In this work, we first develop a memory-bank-based solution~\cite{roth2022towards, wang2023multimodal} that overcomes limitations in modality, network architecture, and the reliance on normal training sample replay. Our approach extracts features from normal training samples and constructs a memory bank to capture their distribution. During inference, test sample features are compared against this memory bank, enabling anomaly detection without requiring a decoder or replaying normal samples. This design ensures broad generalization across different modalities and encoder architectures.

To further address the limitation of requiring prior knowledge of distribution shifts, we freeze the training stage with clean normal data and adapt the model using a small set of target adaptation data. In many scenarios, collecting some unlabeled target-domain data is feasible at low or no cost, enabling model adaptation. Domain adaptation methods~\cite{su2023revisiting, su2022revisiting, liu2021ttt++, liang2021source} leverage such data, with distribution alignment proving particularly effective. By modeling source and target distributions and minimizing their discrepancy, distribution alignment improves generalization with fewer hyperparameter sensitivities compared to self-training or self-supervised learning.

However, existing distribution alignment methods often overlook the presence of anomalies in target data which is common in anomaly detection tasks. Naively aligning distributions without accounting for anomalies can lead to misalignment, as anomalies may dominate the distribution discrepancy. To address this, we draw inspiration from robust distribution alignment techniques in generative modeling~\cite{adler2018banach} and domain adaptation~\cite{courty2016optimal}, formulating the adaptation process as an optimal transport problem. Given target training data, we compute a soft assignment between target samples and memory bank prototypes using optimal transport, e.g., the Sinkhorn distance~\cite{cuturi2013sinkhorn}. To mitigate the risk of assigning anomalies to prototypes, we introduce two key techniques. First, we discretizatize/binarize the Sinkhorn assignment to suppress spurious alignments. We further conduct target domain augmentation to generate multiple views of target samples, increasing robustness to anomalies. These strategies enhance the robustness of Sinkhorn distance, ensuring more reliable adaptation. These designs give rise to our method, \textbf{Ro}bust \textbf{D}istribution \textbf{A}lignment~(\textbf{RoDA}) with an overview illustrated in Fig.~\ref{fig:framework}. Finally, the adapted model is applied for inference on target test data, improving anomaly detection under distribution shifts.

Our contributions are summarized as follows.

\begin{itemize}
    \item We identify the practical challenges of generalizing to OOD target data in industrial anomaly detection and propose a distribution alignment-based optimal transport approach to improve robustness.
    \item We enhance the optimal transport framework for anomaly detection by introducing assignment discretization and target domain data augmentation, mitigating the impact of anomalies.
    \item We conduct evaluations under synthetic distribution shifts across both 2D and 3D data modalities, demonstrating the effectiveness.
\end{itemize}

%% file: sec/2_relatedwork.tex
\section{Related Works}
\noindent\textbf{Anomaly Detection}: 
Anomaly Detection (AD) identifies samples that deviate significantly from the norm. Most AD approaches operate in an unsupervised setting, leveraging different techniques to model normal data~\cite{DeepSVDD18, PMAD23, deng2022anomaly, roth2022towards, deng2022anomaly, xie2023pushing}. One-class classification methods, ~\cite{DeepSVDD18,liu2023simplenet}, represent normal data using support samples. Reconstruction-based methods~\cite{PMAD23, zhang2024realnet} detect anomalies through higher reconstruction errors. Knowledge distillation approaches, such as RD4AD~\cite{deng2022anomaly, tien2023revisiting}, identify anomalies by comparing distilled and original features. Memory bank based methods~\cite{zavrtanik2021draem, roth2022towards} measure deviations in feature space from normal training samples. Among these methods, memory bank based exhibits more versatility to different data modalities and network architectures due to the decoder-free architecture.
% Recently, AD under distribution shifts has gained attention. For example, \cite{cao2023anomaly，kashiani2024roads} enhances reverse distillation~\cite{deng2022anomaly} by augmenting test data to improve generalization. However, these methods assume that test-time shifts resemble training augmentations. In contrast, our work tackles a more practical setting where distribution shifts at test time are substantially different, and normal training samples are unavailable.

\noindent\textbf{Domain Adaptation}: 
Domain adaptation mitigates performance degradation caused by distribution shifts between training and testing data. Traditional methods, such as learning invariant representations~\cite{ganin2015unsupervised} and clustering~\cite{tang2020unsupervised}, rely on both source and target data, which is impractical when source data is restricted due to privacy concerns. This has driven interest in source-free~\cite{liang2020we, liu2021ttt++, yang2021generalized, liang2021source, su2022revisiting, su2023revisiting} and source-light~\cite{bateson2020source, su2023revisiting} adaptation, where models are updated without direct source access to enhance generalization. While effective for classification and segmentation, these methods struggle with anomaly detection. They typically model feature distributions using parametric assumptions (e.g., single Gaussian~\cite{liu2021ttt++} or Gaussian mixtures~\cite{su2022revisiting,su2023revisiting}) and align them via KL-Divergence or moment matching. However, fitting complex industrial data with simple distributions can lead to underfitting~\cite{liu2021ttt++}, and aligning without distinguishing anomalies can blur decision boundaries.
Our method follows a source-light approach, leveraging a lightweight memory bank from source training data. We optimize the optimal transport distance~\cite{cuturi2013sinkhorn} between distributions, a widely studied technique in domain adaptation~\cite{courty2016optimal, lee2019sliced}, and extend it to handle outliers~\cite{balaji2020robust, mukherjee2021outlier}. Our approach improves efficiency and scalability by refining the Sinkhorn distance through discretization and target domain augmentation.

\noindent\textbf{Anomaly Detection under Domain Shift}: 
Anomaly detection under distribution shift has only recently gained attention~\cite{cao2023anomaly, kashiani2024roads}. Existing attempts to address this challenge involved augmenting data during the training stage to enhance the model’s robustness~\cite{cao2023anomaly, kashiani2024roads}, demonstrating effectiveness in both industrial defect detection and natural OOD images. However, these approaches rely on the assumption that training can be modified and that prior knowledge of the distribution shift is available. In this work, we further relax these assumptions by updating the model only during test time upon observing target data, without modifying the training process.
An alternative approach to handling anomaly detection under distribution shifts involves training from scratch using noisy target data~\cite{jiang2022softpatch, chen2022deep, mcintosh2023inter}. These methods incrementally filter out potential anomalies and learn normal patterns from the remaining clean samples. Such methods may struggle to generalize when the noise level in the target distribution is high, limiting their effectiveness in handling severe distribution shifts, as demonstrated by the results in the Appendix.

%% file: sec/3_method.tex
\section{Methodology}

\subsection{Problem Formulation}
We first formally define the task of unsupervised anomaly detection under distribution shift. Without loss of generality, we denote the source domain training data as $\set{D}_s=\{x_i, y_i\}_{i=1\cdots N_s}$ where all samples are defect-free. We further denote the target data as $\set{D}_t=\{x_j,y_j\}_{j=1\cdots N_t}$ where the labels, $y_j$, are not visible. We split the target data into target training, $\set{D}_{tr}$, and target testing, $\set{D}_{te}$, i.e. $\set{D}_t=\set{D}_{tr}\bigcup\set{D}_{te}$. The distributions from which samples are drawn write $\set{D}_s\sim p_s$ and $\set{D}_t\sim p_t$. For anomaly detection purpose, the label only takes a binary value, i.e. $y\in\{0,1\}$ with $1$ indicating anomalous. 

\noindent\textbf{Revisiting Memory-bank Based Anomaly Detection
% PatchCore
}: Following the practice of memory bank based anomaly detection methods~\cite{roth2022towards}, a backbone network extracts features, $z_i=f(x_i;\Theta)\in\mathbbm{R}^{N_p\times D}$, as $N_p$ patches, from input sample $x_i$. A memory bank $\set{M}$ takes an abstraction of source domain training sample patches by sampling a core-set %$\set{C}(\cdot,K)$ 
of size $N_M$ as in Eq.~\ref{eq:coreset}. 
\vspace{-0.2cm}
\begin{equation}\label{eq:coreset}
\begin{split}
    &\set{M}=\arg\min_{\set{M}\subset \set{D}_s}\max_{z_j\in\set{D}_s}\min_{m_i\in \set{M}} ||m_i-z_j||,\\
    &\text{s.t.}\quad|\set{M}|\leq N_M.
\end{split}
\end{equation}

At inference stage, testing sample features are compared against the memory bank to determine anomaly as follows.
\vspace{-0.3cm}
\begin{equation}\label{eq:patchcore_inf}
    s_i = \max_{p\in 1\cdots N_p}\min_{m_k \in \set{M}} ||z_{ip} - m_k||.
\end{equation}
\vspace{-0.2cm}

The above procedure achieves competitive results for industrial defect identification. %Nevertheless, we witness a significant performance drop when testing data experiences a distribution shift, i.e. $p_s\neq p_t$. In this work, we aim to address the distribution shift challenge from a distribution alignment perspective.

\begin{figure}[!tb]
    \centering
    \includegraphics[width=0.99\linewidth]{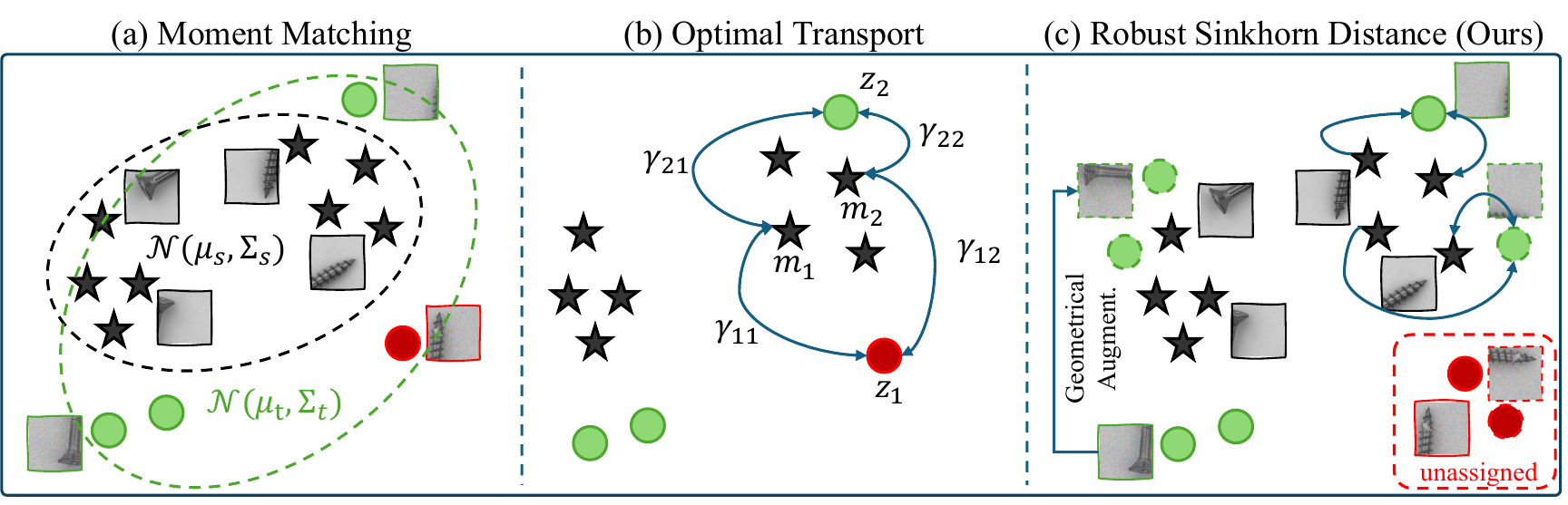}
    \vspace{-0.2cm}
    \caption{Illustration of distribution alignment via moment matching, optimal transport and finally our modified robust sinkhorn distance.}
    \vspace{-0.5cm}
    \label{fig:distshift}
\end{figure}

\subsection{Distribution Alignment for Domain Adaptation}

Under the vanilla memory bank based anomaly detection, we witness a significant performance drop when testing data experiences a distribution shift, i.e. $p_s\neq p_t$. 
The mismatch results in an overall increase of anomaly scores $s_i$. The increment of score is often non-deterministic and inhomogeneous among patches, i.e., anomaly scores for normal and abnormal patches do not simply increase linearly, 
% thus leading to unpredictable changes of anomaly scores between normal and abnormal in the target data,
resulting in unreliable predictions using original PatchCore~\cite{roth2022towards}, which assumes \textit{i.i.d.} test set.
% smaller margin of anomaly scores between normal and abnormal in the testing data, resulting in poorer discrimination between normal and anomalies.

% the increase of  This is caused by the the increased %In this work, we aim to address the distribution shift challenge from a distribution alignment perspective.
% We first identify the underlying reason of why anomaly detection model. %After revisiting the mechanism of memory bank based anomaly detection, we notice the anomaly score for each patch is obtained as the shortest distance to any samples in the memory bank. 
% We provide a 
% Following Eq.~\ref{eq:patchcore_inf}, the anomaly score for each patch is obtained as the shortest distance to any samples in the memory bank. 
% The above way to characterize anomaly score is built upon the assumption that normal sample distribution is consistent between training and testing data. Therefore, a high anomaly score indicates anomaly. This assumption no longer holds true when distribution shift exists as the overall distance between testing patches and memory bank patches are increased, thus diminishing the discriminability between normal and anomalies. 

\noindent\textbf{Distritbuion Alignment via Moment Matching}: Without knowing the target domain distribution shift in advance, recent works proposed distribution alignment to mitigate the distribution shift~\cite{su2022revisiting,liu2021ttt++,su2023revisiting}. %The key insights derived suggest that minimizing the distribution discrepancy between the overall feature distribution of source and target domains could substantially improve the generalization capability. 
Specifically, a parametric distribution, e.g. multi-variate Gaussian distribution~\cite{liu2021ttt++} or mixture of Gaussian distributions~\cite{su2022revisiting,su2023revisiting}, is fitted on both source and target domain, denoted as $p_s(z)$ and $p_t(z)$. A loss function that measures the discrepancy between $p_s(z)$ and $p_t(z)$ is employed. For example, \cite{liu2021ttt++} introduced minimising the L2 distance between mean and covariance between the two Gaussian distributions~(moment matching) for alignment as in Eq.~\ref{eq:KLD}. %A closed-form solution exists and serves as the loss function to optimize on the target domain.

\vspace{-0.2cm}
\begin{equation}\label{eq:KLD}
\min_{\Theta} \mathcal{L}_{DA}, \quad \text{s.t.}\quad
    \mathcal{L}_{DA}=||\mu_s-\mu_t||_2^2 + ||\Sigma_s-\Sigma_t||_F^2.%=D_{KL}(\mathcal{N}(\mu_s,\Sigma_s)||\mathcal{N}(\mu_t,\Sigma_t)).
\end{equation}

Despite the success for classification tasks, we argue that such a vanilla distribution alignment approach is suboptimal for memory bank based anomaly detection task due to the following reasons. As shown in Fig.~\ref{fig:distshift} (a), without knowing the prior information of the feature distribution, fitting arbitrary sample features with a single multi-variate Gaussian distribution is prone to underfitting.
% fitting the model with a single multi-variate Gaussian distribution is prone to underfitting.
A mixture of Gaussian distributions may better fit the complex distribution, however, this requires complicated approximations for KL-Divergence~\cite{hershey2007approximating}. Moreover, target distribution is fitted on target domain contaminated with anomalies. Aligning the target distribution with source distribution will undermine the discrimination between normal and abnormal samples.
% Finally, when the distribution overlap is too small, the gradient of KL-Divergence may be too small, prohibiting gradient-based optimization. Given the above challenge, we resort to a more stable solution to distribution alignment via optimal transport.

\noindent\textbf{Distribution Alignment via Optimal Transport}:
Inspired by the success of distribution based via optimal transport for unsupervised domain adaptation~\cite{courty2017joint,damodaran2018deepjdot}, we propose to use optimal transport~(OT) distance for robust distribution alignment between $\set{M}$ and $\set{D}_{tr}$. Specifically, a cost matrix $C\in\mathbbm{R}^{N_{tp} \times N_M}$ is built between target training patches and source memory bank with $C_{ij}=||z_i-m_j||$ and $N_{tp}=N_t\cdot N_p$. Assuming uniform weight applied to each sample, the optimal transport is formulated as,
\vspace{-0.2cm}
\begin{equation}
\begin{split}
    &\min_{\gamma\geq 0} \sum_i^{N_{tp}}\sum_j^{N_M} \gamma_{ij}C_{ij},\\
    \text{s.t.}\quad &\sum_i\gamma_{ij}=\frac{1}{N_M}, \;\sum_j\gamma_{ij}=\frac{1}{N_{tp}}.
\end{split}
\end{equation}
\vspace{-0.2cm}

% OT solves the following problem with $\Pi(p_t,p_s)$ denoting the joint distribution and $\matr{C}\in\mathbbm{R}^{N_p\times K}$ denoting the pairwise cost. 

% \begin{equation}
% \begin{split}
% OT(p_t,p_s)&=\min_{\gamma\in\Pi(p_t,p_s)}||\gamma\circ \matr{C}||_1,\\
% &\approx \min_{\gamma\in\{0,1\}^{N_p\times K}}||\gamma\circ \matr{C}||_1, \quad \text{s.t.}\quad c_{pk}=||z_{ip}-m_k||_2,\; \sum_{i}\gamma_i = 1
% \end{split}
% \end{equation}

Solving the above problem, through linear programming, is expensive and an efficient algorithm, Sinkhorn distance~\cite{cuturi2013sinkhorn}, exists that can substantially reduce the computation cost. Specifically, an entropy regularization term is added to improve the smoothness and robustness of solution, as in Eq.~\ref{eq:sinkhorn}. An iterative algorithm is employed to solve the problem with details explained in the supplementary. %The Wasserstein distance is then approximated by $\mathcal{L}_{ws}=\sum_{ij} \gamma_{ij}C_{ij}$.

\vspace{-0.5cm}
\begin{equation}\label{eq:sinkhorn}
\begin{split}
    &\min_{\gamma\geq 0} \sum_i^{N_{tp}}\sum_j^{N_M} \gamma_{ij}C_{ij} + \epsilon\sum_i\sum_j \gamma_{ij}\log \gamma_{ij}, \\
    \text{s.t.}\quad &\sum_i\gamma_{ij}=\frac{1}{N_M}, \;\sum_j\gamma_{ij}=\frac{1}{N_{tp}}.
\end{split}
\end{equation}

\noindent\textbf{A Self-Training Perspective}: We further elaborate the distribution alignment from a self-training~(ST) perspective. ST has been demonstrated to be effective for domain adaptation~\cite{su2023revisiting}. The regular routine makes predictions on target domain samples and use most confident ones, a.k.a. pseudo labels, to train network, e.g. optimize cross-entropy loss for classification task. In the realm of anomaly detection, self-training could translate into encouraging target domain patch to be close to the closest patch in the memory bank. Distribution alignment via optimal transport can be seen as discovering a global optimal assignment between target patches and memory bank, as in Fig.~\ref{fig:distshift} (b). The assignment can be seen as the pseudo label and minimizng the earth moving distance is equivalent to using the pseudo label for self-training.

% In the realm of anomaly detection, the self-training loss could be implemented as encouraging normal patches to form tight clusters~\cite{liao2024coft}. This objective could translate into 

\subsection{Robust Sinkhorn Distance}

Solving the optimal transport problem in Eq.~\ref{eq:sinkhorn} yields the assignment $\gamma^*$ for each target sample to source samples. The Sinkhorn distance, $\mathcal{L}_{DA}=\sum_{i}^{N_{tp}}\sum_{j}^{N_M}\gamma^*_{ij}C_{ij}$, could be adopted as the objective to optimize for distribution alignment.
% \begin{equation}
% \mathcal{L}_{DA}=\sum_{i}^{N_{tp}}\sum_{j}^{N_M}\gamma^*_{ij}C_{ij}
% \end{equation}
However, we notice an unresolved issue by directly optimizing the above objective. First, an anomalous patch, indexed by $j^*$, in the target domain are always assigned to source patches in the memory bank due to the constraint $\gamma_{ij^*}\geq 0,\;\sum_{i}\gamma_{ij^*}=\frac{1}{N_M}$. Minimizing the distance between anomalous patches and memory bank patches will inevitably diminish the discriminability. To improve the robust of optimal transport for distribution alignment, we convert the continuous optimal transport assignment into discrete assignment. Fortunately, the discretization may eliminate weak assignments that often appear on anomalous patches in the target domain. Specifically, we follow the rules below to discretize the assignment, resulting in a more robust distribution alignment loss in Eq.~\ref{eq:discrete_sinkhorn}. %We show that the above discretization could substantially reduce the overall assignment between anomalous patches and memory bank patches.

\vspace{-0.5cm}
\begin{equation}
\label{eq:discrete_sinkhorn}
\begin{aligned}
&\quad\min_{\Theta}\mathcal{L}_{\text{DA}}, \quad
\text{s.t.}\quad\mathcal{L}_{\text{DA}} = \sum_{i}^{N_{tp}} \sum_{j}^{N_M} \pi^*_{ij} C_{ij}, \\
&\pi^*_{ij} = 
\begin{cases} 
1 & \text{if } j = \arg \max\limits_j \ \gamma^*_{ij} \text{, or } i = \arg \max\limits_i \ \gamma^*_{ij}, \\ 
0 & \text{otherwise.} 
\end{cases}
\end{aligned}
\end{equation}
\vspace{-0.3cm}

\noindent\textbf{Target Domain Data Augmentation}: We apply a batchwise update strategy to facilitate gradient based update of backbone weights. Wihin each minibatch we further apply data augmentation $\mathcal{T}(x)$ on the target training data to improve the effectiveness distribution alignment, $\tilde{\set{D}_t}=\{\set{T}(x_j)\}_{j=1\cdots N_t}$. The augmentation simulates the normal data variation, e.g. rotation in multiple of $90^\circ$,  etc. Importantly, the augmentation is agnostic to the corruption (distribution shift) on the target domain and is only applied at adaptation stage, in contrast to the training stage augmentation adopted in~\cite{cao2023anomaly, kashiani2024roads}. We attribute the effectiveness of test-time target domain augmentation to the following reasons. First, the augmentation will create a more diverse and smoother distribution. This can help mitigate the impact of outliers by ``diluting" their influence, making the alignment focus on general features rather than outlier-specific characteristics. Moreover, data augmentation can help by incorporating additional noise into the training process in a controlled way, making the model more resilient to noise and outliers in the real world. Fig.~\ref{fig:distshift} (c) has shown the main differences between our robust sinkhorn distance and the previous methods. The positive effect is demonstrated by the reduced discretised assignment, as in Fig.~\ref{fig:tardataaug} (a).

% Directly optimizing approximated wasserstein distance cloud be slow because gradient has to backpropagate through the iterative optimization. We use the approximated assignment and use the assignment to compute the EMD distance

% Directly optimizing the approximated Wassertein distance (Eq.~\ref{eq:wassertein}) for distribution alignment assumes both source and target domain data are free of noisy samples, e.g. anomalies. 

% Anomalies in the target domain could result in signficant contribution to the OT distance and failing distinguishing normal and abnormal samples after alignment. Therefore, we propose an crucial step to further adapt the sinkhorn distance to distribution alignment with outliers.

% An important step of sinkhorn distance measures the forward and backward ...

% Only updated BN layers

\subsection{Overall Algorithm}

Following the practice of common test-time training strategies, we update the batchnorm affine parameters, $\Theta_{bn}$, with distribution alignment loss. We present the overall algorithm of the proposed method in Algo.~\ref{alg:TTTAD}.

\begin{algorithm}
\caption{Domain Adaptation for Anomaly Detection}
\label{alg:TTTAD}
\begin{spacing}{1.0}  % 调整行距
\begin{algorithmic}[1]  % The number [1] means each line will be numbered
    \State \textbf{Input:} Pretrained memory bank $\set{M}$, target training data $\set{D}_{tr}$, target testing data $\set{D}_{te}$, initial encoder network $\Theta$
    
    \State \textbf{Output:} Anomaly scores $\{s_i\}$

    \noindent\textcolor{gray}{\# Domain Adaptation on target training data}
    \For{$\set{B}_t \subset \set{D}_{tr}$} \textcolor{gray}{\# Collect one minibatch $\set{B}_t$}

        Augment target minibatch $\tilde{\set{B}}_t=\set{T}(\set{B}_t)$

        Compute cost matrix $C\in\mathbbm{R}^{|\set{D}_s|\times |\tilde{\set{B}}_t|}$
        
        Solve optimal transport plan $\gamma^*$ by Eq.~\ref{eq:sinkhorn}
        
        Discretize assignment $\pi^*$ by Eq.~\ref{eq:discrete_sinkhorn}
        
        Update model $\Theta_{bn}=\Theta_{bn}-\alpha\frac{\nabla \mathcal{L}_{DA}}{\Theta_{bn}}$
    \EndFor
    
\noindent\textcolor{gray}{\# Evaluate on target testing data}
    
    \For{$x_i \in \set{D}_{te}$}

        Encode feature with updated model $z_i=f(x_i;\Theta^*)$
        
        Per sample anomaly score $s_i = \max\limits_{p\in 1\cdots N_p}\min\limits_{m_k \in \set{M}} ||z_{ip} - m_k|| _2$
        
    \EndFor

    \State \textbf{return} $s_i$  % Return the predicted result

\end{algorithmic}
\end{spacing}
\end{algorithm}

%% file: sec/4_experiments.tex
\section{Experiments}

% We first introduce the datasets, evaluation protocols and competing methods. We then analyze the results on three industrial anomaly detection datasets and present qualitative results.
\subsection{Experiment Settings}
\noindent\textbf{Dataset}: 
We evaluate our method on two widely-used 2D industrial anomaly detection datasets, \textbf{MVTec} \cite{bergmann2019mvtec} and \textbf{RealIAD} \cite{wang2024real}, as well as on a 3D dataset, \textbf{MVTec 3D} \cite{bergmann2021mvtec}. 
%\textbf{MVTec} is the most commonly used benchmark for 2D industrial anomaly detection, comprising 15 object categories, with 60-300 normal samples for training and 30-400 normal and anomalous samples for testing. \textbf{RealIAD} is a newly introduced industrial dataset with 30 object categories, each captured from five different viewpoints. We follow the single-view experiment setup, utilizing only the top-view images. Due to the high resolution of the original images (over 3,000×5,000 pixels), which imposes significant computational demands, we use a downsampled version with a resolution of 1,024×1,024. Illustration of the two 2D dataset is shown in \ref{fig:samples_small}. \textbf{MVTec 3D} consists of 3D scans that include both geometric surface data and RGB information. The dataset comprises 10 object categories, with over 200 normal images for training and more than 100 images for testing per category.
% \begin{figure}
%     \centering
%     \includegraphics[width=1\linewidth]{ICLR 2025/figure/samples_small.pdf}
%     \vspace{-0.5cm}
%     \caption{Illustrations of the distribution shift on MVTec and RealIAD dataset. Severity level of all corruptions are set to 5. It can be observed that the difficulty varies significantly across different corruptions, with contrast being the most challenging. More examples are shown in Appendix Fig. \ref{fig:MVTec_samples} \ref{fig:IAD_samples}}.
%     \label{fig:samples_small}
% \end{figure}
\textbf{MVTec} is the most commonly used benchmark for 2D industrial anomaly detection, comprising 15 object categories, with 60-300 normal samples for training and 30-400 normal and anomalous samples for testing. \textbf{RealIAD} is a newly introduced industrial dataset with 30 object categories, each captured from five different viewpoints. We follow the single-view experiment setup, utilizing only the top-view images. Due to the high resolution of the original images (over 3,000×5,000 pixels), which imposes significant computational demands, we use a downsampled version with a resolution of 1,024×1,024. 
% Illustration of the two 2D dataset is shown in \ref{fig:samples_small}. 
\textbf{MVTec 3D} consists of 3D scans that include both geometric surface data and RGB information. The dataset comprises 10 object categories, with over 200 normal images for training and more than 100 images for testing per category.

% We evaluate our methods on two widely-used 2D industrial anomaly detection datasets, \textbf{MVTec} \cite{bergmann2019mvtec} and \textbf{Real-IAD} \cite{wang2024real}, and a 3D dataset, \textbf{MVTec3D} \cite{bergmann2021mvtec}. \textbf{MVTec} is the most commonly used benchmark dataset for 2D industrial anomaly detection, consisting of 15 object categories, with 60-300 normal samples for training and 30-400 normal and anomalous samples for testing. \textbf{Real-IAD} is a newly introduced industrial dataset with 30 object categories, each captured from 5 different viewpoints. We follow the single-view experiment setup, using only the top-view images. Due to the high resolution of the original images (over 3,000×5,000 pixels), which imposes significant computational demands, we use a downsampled version with 1,024×1,024 resolution. 
% \textbf{MVTec 3D} comprises 3D scans that include both geometric surface data and RGB information. The dataset consists of 10 object categories, with over 200 normal images for training and more than 100 images for testing per category. %Examples of above datasets can be found in Appendix Figure \ref{fig:MVTec_samples}.

\noindent\textbf{Evaluation Protocol}: 
We simulate commonly seen distribution shift in the testing set to evaluate the generalization robustness. 
For the 2D datasets, MVTec and RealIAD, we follow the corruption generation process described in \cite{hendrycks2019benchmarking}, applying four common corruptions, including Gaussian Noise, Defocus Blur, Contrast, and Brightness, with a severity level of 5 to synthesize distribution-shifted data. In the 3D dataset, MVTec~3D, we simulate natural distribution shifts by randomly adding Gaussian noise $n \sim \mathcal{N}(0, [1e-6]^2)$ to the images. 

The distribution shifted test set is then split into two parts: 20\% for model adaptation and 80\% for final evaluation. To ensure a fair comparison, all baseline methods are evaluated on the same 80\% test set. 
We assess the performance using the area under the ROC curve (AUROC), treating anomalies as the positive class for both anomaly detection and segmentation tasks, following the standard protocol~\cite{bergmann2019mvtec}.
The remaining experimental settings are provided in the Appendix.

% Specifically, our method is applicable to both 2D and 3D scenarios, where we adopt different anomaly detection backbones for each setting: PatchCore for 2D data and Point-MAE for 3D data. We follow the standard setting of the two models. \\
% For both 2D datasets, we adopt the corruption generation process from \cite{hendrycks2019benchmarking} and apply three common corruptions: Gaussian Noise, Defocus Blur and Brightness, with serverity of 5, to create distribution-shifted data. For the 3D dataset, we randomly add Gaussian Noise $n \sim \mathcal{N}(0, [1e-6]^2)$ on the images to simulate the natural distribution shift.
% The area under the ROC (AUROC) is used to assess performance, treating anomalies as the positive class, for both anomaly detection and segmentation.

\noindent\textbf{Competing Methods}: 
We compare against several baseline methods, covering several state-of-art industrial AD methods, and three domain adaptation AD methods. These 2D industrial AD methods including reconstruction-based approaches (\textbf{ViTAD}~\cite{vitad}), embedding-based methods (\textbf{CFLOW-AD}~\cite{gudovskiy2022cflow}), and knowledge distillation methods (\textbf{KDAD}~\cite{salehi2021multiresolution} and \textbf{RD4AD}~\cite{deng2022anomaly}). We also evaluated on a unified model (\textbf{UnIAD}~\cite{you2022unifiedmodelmulticlassanomaly}), and the effective memory-bank-based method (\textbf{PatchCore}~\cite{roth2022towards}). 
We further incorporate domain adaptation methods for anomaly detection task. In specific, we evaluated two distribution alignment based approaches, \textbf{TTT++}~\cite{liu2021ttt++} and \textbf{TTAC}~\cite{su2022revisiting}, on top of PatchCore. Additionally, the state-of-the-art domain adaptation method for anomaly detection, \textbf{GNL}~\cite{cao2023anomaly}, are benchmarked. We allow GNL to be trained with default data augmentation. For 3D anomaly detection, we also benchmark several hand-crafted features implemented by \cite{bergmann2021mvtec}, FPFH~\cite{horwitz2022empirical} and M3DM~\cite{wang2023multimodal}. Finally, we evaluate our proposed method, \textbf{RoDA}, on all datasets.

% For the domain adaptation methods, we incorporate two prominent test time training method TTT++ \cite{NEURIPS2021_b618c321} and TTAC \cite{su2022revisiting} with PatchCore, and evaluate on the prior work GNL \cite{cao2023anomaly}. 
% In the 3D dataset experiments, we explore the methods proposed in \cite{bergmann2021mvtec} and PointMAE \cite{pang2022maskedautoencoderspointcloud}.

\subsection{Distribution Shifted Anomaly Detection}

We first present the anomaly detection results, averaged across all object classes, on both the MVTec and RealIAD datasets in Tab.~\ref{tab:AnomalyDetect2D}. More detailed results for per-class AUROC are deferred to the Appendix. From the results, we make the following key observations. \textbf{i)} State-of-the-art anomaly detection methods struggle significantly under distribution shifts, as evidenced by the performance gap when tested on clean versus corrupted target data. For instance, PatchCore shows a performance drop of 18.58\% when exposed to Gaussian noise on the MVTec dataset. This highlights the vulnerability of these methods to out-of-distribution (OOD) scenarios.
\textbf{ii)} Domain adaptation methods (e.g., TTT++ and TTAC), despite showing strong performance on classification tasks, fall short on anomaly detection. TTAC generally underperforms compared to PatchCore, and while TTT++ improves on MVTec, it struggles on the more challenging RealAd dataset. This can be attributed to their reliance on modeling complex distributions with a single Gaussian, which is underfitting.
\textbf{iii)} In contrast, RoDA, which leverages distribution alignment via optimal transport, demonstrates superior performance in 3 out of 4 types of corruptions, with the sole exception being the ``Contrast" corruption. Notably, under the ``Defocus Blur" and ``Brightness" corruptions on the MVTec dataset, RoDA's performance is only 2\% behind the results on source domain. These findings underscore the importance of a well-calibrated distribution strategy for robust anomaly detection.
\textbf{iv)} Lastly, we observe that GNL significantly outperforms all competing methods under the ``Contrast" corruption scenario. Upon further investigation, we discovered that GNL employs an ``AutoContrast" augmentation during training on source domain, which inadvertently provides prior knowledge of the target data distribution. This advantage highlights the importance of evaluating methods under consistent and unbiased conditions.

\begin{table*}[htbp]
  \centering
  \vspace{-0.3cm}
  \caption{{Results of anomaly detection on MVTec and RealIAD datasets. We report the mAUROC(\%) averaged across all classes. ``Clean'' refers to the results on testing samples without distribution shift.}}
  \vspace{-0.2cm}
  
\setlength{\tabcolsep}{3pt}
\resizebox{0.99\textwidth}{!}{
    \begin{tabular}{c|l|c|c|c|c|c|c|c|c|c|c|c|c}
    % \hline
    \toprule[1.2pt]
           \multicolumn{1}{c}{} & 
           & \multicolumn{6}{c|}{MVTec}     & \multicolumn{6}{c}{RealIAD} \\
          \cmidrule(lr){3-8}\cmidrule(lr){9-14} \multicolumn{1}{c}{} & & Clean & Gauss.~Noise & Defoc.~Blur & Bright. & Contrast & Mean & Clean & Gauss.~Noise & Defoc.~Blur & Bright. & Contrast & Mean \\
          % \hline
    \midrule
    \multirow{6}{*}{\rotatebox[origin=c]{90}{\footnotesize{w/o adapt.}}}
    & ViTAD~\cite{vitad} & 98.30 & 64.91 & 80.15	& 65.67	& 54.07 & 66.20 & 82.70 & 52.50 &	73.46 & 60.90 & 57.21 & 61.02 \\
    & KDAD~\cite{salehi2021multiresolution} & 87.74  & 72.18 & 79.33 & 70.98 & 47.65 & 67.54 & 80.23 & 41.43& 31.15 & 38.36 & 46.80 & 39.44 \\
    & RD4AD~\cite{deng2022anomaly} & 98.50 & 78.09	& 89.45 & 86.55 & 62.02 & 79.03 & 86.17 & 56.57 & 79.55 & 63.65 & 57.56 & 64.33 \\
    & UnIAD~\cite{you2022unifiedmodelmulticlassanomaly} & 92.50 & 82.00	& 91.00	& 85.33 &	60.13  & 79.61  & 83.10 & 64.21 &	78.81 &	69.41 & 54.07 & 66.63 \\
    & CFLOW-AD~\cite{gudovskiy2022cflow} & 91.55 & 60.69 & 60.23 & 55.88	& 51.45 & 57.06 & 77.00 & 56.45 & 62.59 & 56.38 & 53.24 & 57.17\\
    & PatchCore~\cite{roth2022towards} & 98.81 & 80.23 & 91.22 & 91.91 & 60.89 & 81.06  & 90.35 & 60.27 & 76.75 & 62.60 & 49.65 & 62.32\\
    \midrule
    \multirow{4}{*}{\rotatebox[origin=c]{90}{\footnotesize{w/ adapt.}}}
    & TTAC~\cite{su2022revisiting} & 98.81 & 58.99 & 81.49 & 57.43 & 56.49 & 63.60 & 90.35 & 52.11 & 50.73 & 44.53 & 50.52 & 49.47 \\
    & TTT++~\cite{liu2021ttt++} & 98.81 & 84.06 & 94.61 & 93.32 & 71.52 & 85.88 & 90.35 & 62.96 & 72.09 & 62.68 & 57.21 & 63.74 \\
    & GNL~\cite{cao2023anomaly}  &  97.99 & 83.26 & 95.18 & 93.55 & \textbf{86.50} & 89.62 & 83.44  & 68.80 & 73.78 & 68.90 & \textbf{62.26} & 68.44 \\
    % \hline
    % \midrule
    & RoDA~(\textbf{Ours})  &  \bf 98.81 & \bf 87.92 & \bf 96.56 & \bf 96.44 & 84.45  & \bf 91.34 & \bf 90.35 & \bf 72.48 & \bf 80.11 & \bf 71.17 & 60.22 & \bf 71.00 \\
    % \hline
    \bottomrule[1.2pt]
    \end{tabular}%
    }
  \label{tab:AnomalyDetect2D}%
\end{table*}%

In addition to the experiments on 2D data, we also evaluated our method on 3D data by introducing Gaussian noise as a form of corruption. The results are presented in Tab.~\ref{tab:3dMvtecResults}. Compared to standard 2D corruptions, adding Gaussian noise to 3D data introduces a greater challenge. The geometric structures in 3D data are particularly sensitive to noise, as it disrupts fine details and depth information, both of which are crucial for effective 3D anomaly detection. Despite these challenges, our method demonstrates resilience, achieving a notable performance improvement of 14.21\% on PointMAE. Compared with other adaptation methods, our approach achieves a 2.33\% improvement over the second-best TTT++ adaptation. This suggests that our approach is capable of effectively managing the complexities introduced by noise in 3D data, maintaining robust anomaly detection capabilities.

% In addition to the experiments on 2D data, we also conduct experiments on 3D data by adding Gaussian noise to simulate a corruption. The results are shown in Table \ref{tab:3dMvtecResults}. Compares to standard 2D corruptions, adding Gaussian noise to 3D data presents a greater challenge. The geometric structures of 3D data is more sensitive to noise, as it disrupts the fine details and depth information that are critical for 3D anomaly detection. Despite this, our method continues to demonstrate robust increase of 1.5\% on PointMAE, effectively handling the noise. The per-class results for all experiments are reported in Appendix Table \ref{tab:MVTecDetectPerclass} \ref{tab:RealIADDetectPerclass} \ref{tab:MVTec3DDetectPerclass}.

% Table generated by Excel2LaTeX from sheet 'Sheet6'
\begin{table*}[htbp]
  \centering
  % \vspace{-0.3cm}
  \caption{Results of anomaly detection on MVTec-3D with per class AUROC(\%)}
  \vspace{-0.2cm}
   \resizebox{0.88\linewidth}{!}{
    \begin{tabular}{c|l|c|c|c|c|c|c|c|c|c|c|r}
    % \hline
    \toprule[1.2pt]
          \multicolumn{1}{c}{} & & \multicolumn{1}{l|}{Bagel} & \multicolumn{1}{l|}{CableGland} & \multicolumn{1}{l|}{Carrot} & \multicolumn{1}{l|}{Cookie} & \multicolumn{1}{l|}{Dowel} & \multicolumn{1}{l|}{Foam} & \multicolumn{1}{l|}{Peach} & \multicolumn{1}{l|}{Potato} & \multicolumn{1}{l|}{Rope} & \multicolumn{1}{l|}{Tire} & \multicolumn{1}{l}{Mean} \\
    % \hline
    \midrule
    \multirow{9}{*}{\rotatebox[origin=c]{90}{\footnotesize{w/o adapt.}}}
    & Depth GAN~\cite{bergmann2021mvtec} & 47.5  & 24.0    & 49.1  & 45.9  & 37.4  & 36.8  & 32.4  & 37.0    & 35.1  & 36.5  & 38.17 \\
    & Depth AE~\cite{bergmann2021mvtec} & 33.4  & 38.6  & 43.3  & 47.9  & 40.7  & 32.3  & 42.9  & 41.6  & 41.2  & 38.3  & 40.02 \\
    & Depth VM~\cite{bergmann2021mvtec} & 36.7  & 32.2  & 37.4  & 44.6  & 40.4  & 29.2  & 38.7  & 29.5  & 45.3  & 39.7  & 37.37 \\
    & Depth PatchCore~\cite{roth2022towards} & 75.8  & 53.8  & 64.3  & 75.5  & 44.6  & 48.4  & 40.8  & 50.7  & 56.5  & 56.6  & 56.70 \\
    & Raw~(in BTF)~\cite{bergmann2021mvtec} & 58.4  & 49.8  & 44.8  & 45.7  & 50.2  & 33.2  & 24.7  & 31.1  & 44.6  & 50.4  & 43.29 \\
    & HoG~(in BTF)~\cite{bergmann2021mvtec} & 61.2  & 57.2  & 33.0    & 56.9  & 51.1  & 41.8  & 38.4  & 69.2  & 50.0    & 60.6  & 51.94 \\
    & SIFT~(in BTF)~\cite{bergmann2021mvtec} & 46.1  & 42.3  & 44.1  & 46.6  & 38.5  & 41.9  & 33.4  & 55.7  & 62.4  & 56.4  & 46.74 \\
    & FPFH~\cite{horwitz2022empirical}  & 49.4  & 48.0    & 54.8  & 37.0    & 38.8  & 38.7  & 36.5  & 50.7  & 51.9  & 49.8  & 45.56 \\
    & M3DM~\cite{wang2023multimodal} & 74.1  & 51.6  & 73.2  & 83.2  & 59.9  &  58.6  & 30.0    & 76.3  &  86.8  &  70.8  & 66.45 \\
    % \hline
    \midrule
    \multirow{3}{*}{\rotatebox[origin=c]{90}{\footnotesize{w/ adapt.}}}
    & TTAC~\cite{su2022revisiting} &  68.1 &	65.9 &	90.8 &	87.0 &	74.0 &	52.8  & 60.5 &	81.3  &	\bf 84.9 &	41.6 &	70.69 \\
    & TTT++~\cite{liu2021ttt++} & 91.2 &	\bf 71.4 &	\bf 93.4 & 88.9 &	75.1 &	65.6 &	66.4	 & \bf 86.7 &	82.2 &	62.4 & 78.33 \\
    & RoDA~(\textbf{Ours})  & \bf 92.2 &	69.2 &	92.6	 & \bf 94.5 &	\bf 82.1 &	\bf 70.6	 & \bf 73.4 &	83.2 &	82.6 & \bf 66.2  &	\bf 80.66\\
    % \hline
    \bottomrule[1.2pt]
    \end{tabular}%
    }
  \label{tab:3dMvtecResults}%
\end{table*}%

% % Table generated by Excel2LaTeX from sheet 'Sheet6'
% \begin{table}[htbp]
%   \centering
%   \caption{Results on MVTec-3D dataset. Averaged AUROC(\%) across all classes.}
%     \begin{tabular}{l|c}
%     \hline
%           & \multicolumn{1}{l}{3D Gaussian Noise} \\
%           \hline
%     depth GAN~\cite{bergmann2021mvtec} & 38.17 \\
%     depth AE~\cite{bergmann2021mvtec} & 40.02 \\
%     depth VM~\cite{bergmann2021mvtec} & 37.37 \\
%     depth PatchCore~\cite{roth2022towards} & 56.70 \\
%     raw(in BTF)~\cite{bergmann2021mvtec} & 43.29 \\
%     HoG(in BTF)~\cite{bergmann2021mvtec} & 51.94 \\
%     SIFT(in BTF)~\cite{bergmann2021mvtec} & 46.74 \\
%     FPFH~\cite{horwitz2022empirical}   & 45.56 \\
%     PointMAE~\cite{wang2023multimodal} & 66.45 \\
%     \hline
%     RoDA (\textbf{Ours}) & \textbf{67.96} \\
%     \hline
%     \end{tabular}%
%     % \vspace{-0.5cm}
%   \label{tab:3dMvtecResults}%
% \end{table}%

\subsection{Distribution Shifted Anomaly Segmentation}

We also evaluate the anomaly segmentation performance on the MVTec and RealIAD dataset, with the results summarized in Tab.~\ref{tab:MVTecSeg}. Detailed results for each class are deferred to the Appendix. For a fair comparison, we include only those methods that provide segmentation solutions in their original papers. As shown in the table, our method consistently achieves superior AUROC across all types of corruption in the segmentation task. Notably, it surpasses all baseline methods across different corruptions on both dataset. While TTT++ also performs well under Defocus Blur on MVTec dataset, our method maintains an advantage. Moreover, under Contrast on MVTec dataset, our approach outperforms TTT++ by significant margins of 9.16\%. On the RealIAD dataset, our method further shows a significant lead in the all corruption types. 

% We also evaluate the anomaly segmentation performance on the MVTec dataset, as shown in Table \ref{tab:MVTecSeg}. For fair comparison, we only validate on methods that provide segmentation solutions in their original paper.\\
% As shown in the table, our method achieves consistently superior AUROC across all corruption types in the segmentation task on the MVTec dataset. Specifically, it outperforms all baselines with scores of 94.13\%, 96.53\%, and 96.44\% under Gaussian Noise, Defocus Blur, and Brightness corruption, respectively. When compared to RD4AD, which also performs well on Defocus Blur, our method still maintains a lead, and RD4AD falls short under Brightness and Gaussian Noise, where our method outperforms it by 7.25\% and 5.61\%, respectively.

% The per-class results can be found in Appendix Table \ref{tab:MVTecSeg_perclass}.

% Table generated by Excel2LaTeX from sheet 'Sheet7'
\begin{table*}[htbp]
  \centering
  \vspace{-0.4cm}
  \caption{Anomaly segmentation results on MVTec and RealIAD datasets. We report the pixel-level P-mAUROC(\%) across all classes.}
  \vspace{-0.3cm}
  \setlength{\tabcolsep}{3pt}
  \resizebox{0.99\textwidth}{!}{
    \begin{tabular}{c|l|c|c|c|c|c|c|c|c|c|c|c|c}
    % \hline
    \toprule[1.2pt]
          \multicolumn{1}{c}{} &  & \multicolumn{6}{c}{MVTec}     & \multicolumn{6}{|c}{RealIAD} \\
          % \hline
    \cmidrule(lr){3-8}
    \cmidrule(lr){9-14}  
        \multicolumn{1}{c}{} &  & Clean & Gauss. Noise & Defoc. Blur & Brightness & Contrast & Mean &  Clean & Gauss. Noise & Defoc. Blur & Brightness & Contrast & Mean \\
        % \hline
    \midrule
    \multirow{4}{*}{\rotatebox[origin=c]{90}{\footnotesize{w/o adapt.}}}
    & CFLOW-AD~\cite{gudovskiy2022cflow} & 95.65 & 70.40 & 78.87 & 75.55 & 52.90 & 69.43  & 88.60  & 71.30 & 91.08 & 85.05 & 72.21 & 79.91 \\
    & UnIAD~\cite{you2022unifiedmodelmulticlassanomaly} & 95.70 & 71.04 & 87.37 & 90.74 & 72.62 & 80.44  &  86.00 & 88.12 & 96.15 & 90.20 & 80.66 & 88.78 \\
    & RD4AD~\cite{deng2022anomaly} & 97.80 &  86.99 & 93.89 & 91.64 & 77.69 & 87.55 &  89.22 & 62.51 & 93.63 & 78.31 & 79.98 & 78.61 \\
    & PatchCore~\cite{roth2022towards} & 98.34  & 88.89 & 94.30 & 92.77 & 71.09  & 86.76 &   98.10  & 76.39 & 96.61 & 83.03 & 74.08 & 82.53\\
    \midrule
    \multirow{3}{*}{\rotatebox[origin=c]{90}{\footnotesize{w/ adapt.}}}
    & TTAC~\cite{su2022revisiting}  & 98.34 & 57.14 & 82.57 & 40.96 & 54.84 & 58.88 & 98.10   & 52.90 & 45.91 & 62.40 & 50.18 & 52.85 \\
    & TTT++~\cite{liu2021ttt++} & 98.34 &  91.09 & 96.69 & 94.61 & 79.79 & 90.55 &  98.10  & 75.35 & 90.22 & 78.67 & 68.16 & 78.10 \\
    % \hline
    & RoDA \textbf{(Ours)} & \bf 98.34 & \bf 92.26 & \bf 97.05 & \bf 96.47 & \bf 88.95 & \bf 93.68 & \bf 98.10  & \bf 93.14 & \bf 97.81 & \bf 91.85 & \bf 84.19 & \bf 91.75\\
    % \hline
    \bottomrule[1.2pt]
    \end{tabular}%
    }
  \label{tab:MVTecSeg}%
  \vspace{-0.2cm}
\end{table*}%

% \FloatBarrier  % 在图片之前加上这个指令，确保图片浮动到合适位置
% \begin{wrapfigure}{r}{0.65\textwidth}
% \vspace{-0.5cm}
% \centering
% \includegraphics[width=0.99\linewidth]{figure/heatmap_v6.pdf}
% \vspace{-0.3cm}
%     \caption{Qualitative results for anomaly segmentation. We present results for PatchCore without adaptation~(w/o Adapt), RD4AD, TTAD~(Ours) and predictions on clean testing sample as upperbound (Source Pred.). TTAD consistently improves anomaly localization compared to the baseline (w/o Adapt), sometimes even approaching the upperbound.\vspace{-0.5cm}}
%     \label{fig:heatmap}
% \end{wrapfigure}

\begin{figure}
% \vspace{-0.5cm}
\centering
\includegraphics[width=0.99\linewidth]{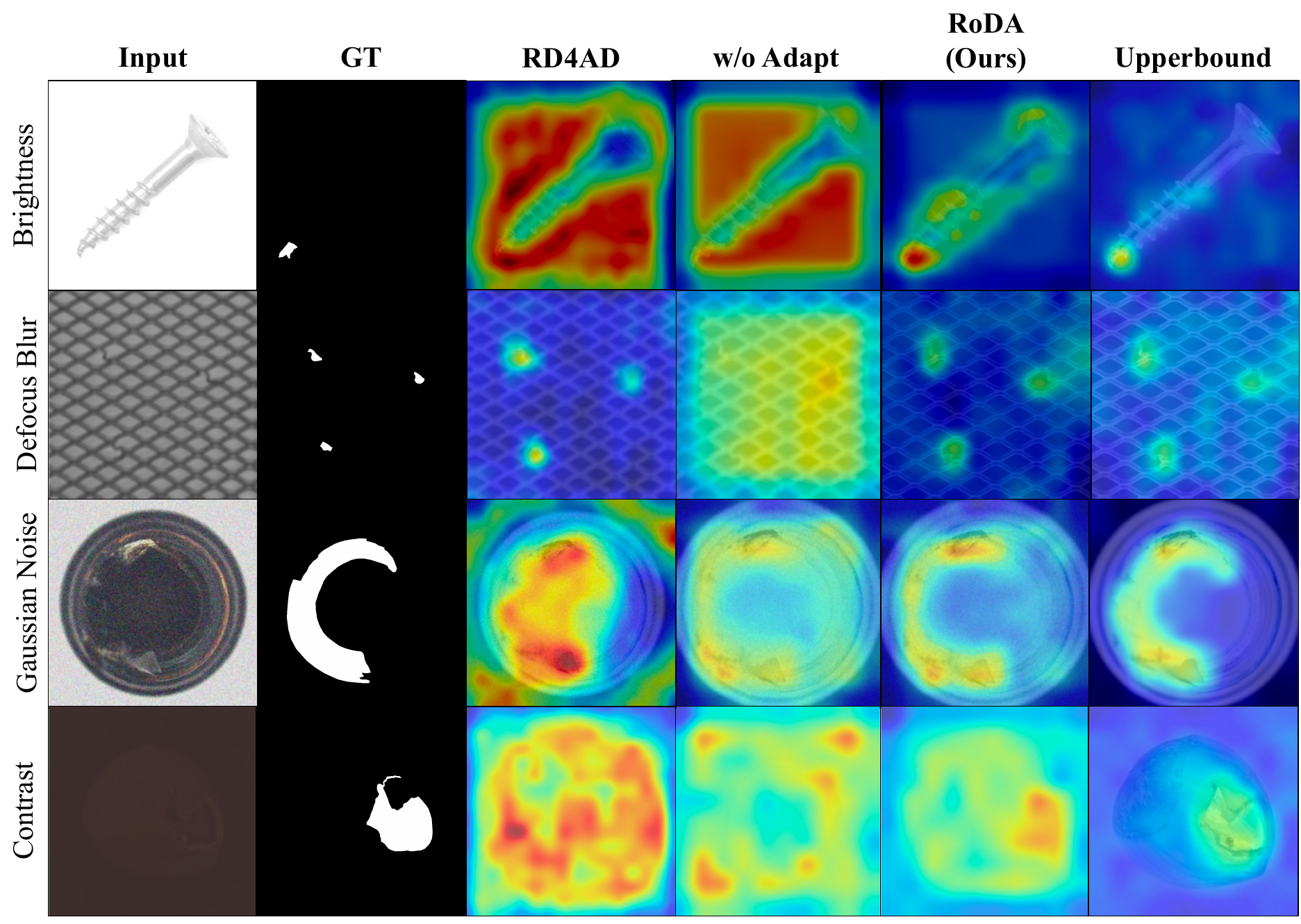}
\vspace{-0.3cm}
    \caption{Qualitative results for anomaly segmentation. We present results for PatchCore without adaptation~(w/o Adapt), RD4AD, RoDA~(Ours) and predictions on original distribution testing sample as Upperbound. RoDA consistently improves anomaly localization compared to the baseline (w/o Adapt), even approaching the Upperbound in some cases.\vspace{-0.5cm}}
    \label{fig:heatmap}
\end{figure}

\noindent\textbf{Qualitative Results}: We provide a qualitative comparison of anomaly segmentation results, as shown in Fig.~\ref{fig:heatmap}. We compare our method with PatchCore without adaptation and the second-best overall adaptation-free baseline, RD4AD. The source predictions serve as reference upperbound. RD4AD performs well under simpler corruptions like Defocus Blur, achieving relatively accurate anomaly localization. However, under more challenging corruptions such as Brightness and Contrast, it tends to misidentify the entire background or object as the anomaly area. In contrast, our method shows a significant improvement compared to the no-adaptation model, which lacks segmentation capability, and demonstrate a strong and consistent performance which is closely approaching the upper bound results on clean samples.

% \begin{figure}[!h]
%     \centering
%     % \vspace{-1cm}
%     \includegraphics[width=1.0\linewidth]{ICLR 2025/figure/heatmap_v4.pdf}
%     \caption{The figure demonstrates a qualitative comparison of anomaly segmentation results under different corruptions across three methods: without adaptation (w/o Adapt), RD4AD \cite{deng2022anomaly}, and our proposed Test-Time Adaptation (TTAD). Additionally, predictions on corruption-free data are shown as the upper bound (Source Pred.). The ground truth (GT) column displays the actual anomaly regions. TTAD consistently improves anomaly localization compared to the baseline (w/o Adapt), closely approaching the performance of the upper bound provided by Source Pred.}
%     \label{fig:heatmap}
% \end{figure}

% \begin{table}[htbp]
%   \centering
%   \caption{Anomaly segmentation results on MVTec dataset. Averaged AUROC(\%) across all classes.}
%     \begin{tabular}{l|c|c|c|c}
%     \hline
%           & \multicolumn{1}{l}{Gaussian Noise} & \multicolumn{1}{|l|}{Defocus Blur} & \multicolumn{1}{|l}{Brightness} &
%           \multicolumn{1}{|l}{Contrast}\\
%     \hline
%     RD4AD & 86.88 & 96.52 & 90.83 \\
%     UnIAD & 70.02 & 90.86 & 87.04 \\
%     CFLOW-AD & 70.38 & 79.19 & 75.87 \\
%     PatchCore & 87.00   & 93.34 & 91.39 \\
%     TTT++ & 80.31 & 79.21 & 79.20 \\
%     TTAC  & 56.77 & 38.43 & 82.28 \\
%     \hline
%     \textbf{Ours}  & \textbf{94.13} & \textbf{96.53} & \textbf{96.44} \\
%     \hline
%     \end{tabular}%
%   \label{tab:MVTecSeg}%
% \end{table}%

% Table generated by Excel2LaTeX from sheet 'Sheet6'
\begin{table*}[htbp]
  \centering
  \caption{Ablation study on MVTec dataset. We report anomaly detection and segmentation AUROC averaged over all classes~(mAUROC \& P-mAUROC).}
  \vspace{-0.3cm}
  \setlength{\tabcolsep}{2pt} % default is 6pt
  \resizebox{0.95\textwidth}{!}{
    \begin{tabular}{c|c|c|cc|cc|cc|cc}
    \toprule[1.2pt]
    \multirow{2.5}{*}{\shortstack{Distribution \\ Alignment}} & \multirow{2.5}{*}{Assignment} & \multirow{2.5}{*}{Target Data Aug.}    & \multicolumn{2}{c|}{Gaussian Noise} & \multicolumn{2}{c|}{Defocus Blur} & \multicolumn{2}{c|}{Brightness} & \multicolumn{2}{c}{Contrast} \\
    % \hline
    % \midrule
    \cmidrule(lr){4-5}
    \cmidrule(lr){6-7}
    \cmidrule(lr){8-9}
    \cmidrule(lr){10-11}
    & & & mAUROC & P-mAUROC & mAUROC & P-mAUROC & mAUROC & P-mAUROC & mAUROC & P-mAUROC \\
    % \hline
    \midrule
    -     & -     & -     & 80.23 & 88.89    & 91.22 & 94.30 & 91.91 & 92.77 & 60.89 &	71.09 \\
    KL-Diver. & -     & -     & 58.99 & 57.14 & 81.49 & 82.57 & 57.43 & 40.96 & 56.49 & 54.84 \\
    Moment Matching    & -     & -     & 84.06 & 91.09 & 94.61 & 96.69 & 93.32 & 94.61 & 71.52 & 79.79 \\
    Optimal Transport & Hun. Method & -     & 82.95 & 88.77 & 93.13 & 95.60 & 93.75 & 94.23 & 72.80 & 80.61 \\
    Optimal Transport & Continuous & -     &  82.51 & 89.10 & 92.55 & 95.39 & 94.16 & 95.61 & 76.01 & 83.21 \\
    Optimal Transport & Discrete & -     & 84.40 & 91.50 & 94.62 & 96.01 & 94.43 & 95.71 & 76.63 & 82.95 \\
    Optimal Transport & Discrete & Direct Copy & 84.47 & 91.55 & 94.64 & 96.04 & 94.41 & 95.65 & 76.65 & 83.01 \\
    Optimal Transport & Discrete & Data Augment. & 
    \bf 87.92 & \bf 92.26 & \bf 96.56 & \bf 97.05 & \bf 96.44 & \bf 96.47 & \bf 84.45 & \bf 88.95 \\
    \bottomrule[1.2pt]
    \end{tabular}%
    }
  \label{tab:ablation}%
    \vspace{-0.3cm}
\end{table*}%

% \begin{table*}[htbp]
%   \centering
%   \caption{Ablation study on MVTec dataset. We explore the variations of the proposed method by employing different assignment and data processing techniques for target distribution data. We report anomaly detection and segmentation AUROC averaged over all classes~(mAUROC \& P-mAUROC).}
%   \setlength{\tabcolsep}{2pt} % default is 6pt
%   \resizebox{\textwidth}{!}{
%     \begin{tabular}{c|c|cc|cc|cc|cc}
%     \toprule[1.2pt]
%     \multirow{2.5}{*}{Assignment} & \multirow{2.5}{*}{Target Data Aug.}    & \multicolumn{2}{c|}{Gaussian Noise} & \multicolumn{2}{c|}{Defocus Blur} & \multicolumn{2}{c|}{Brightness} & \multicolumn{2}{c}{Contrast} \\
%     % \hline
%     % \midrule
%     \cmidrule(lr){3-4}
%     \cmidrule(lr){5-6}
%     \cmidrule(lr){7-8}
%     \cmidrule(lr){9-10}
%     & & mAUROC & P-mAUROC & mAUROC & P-mAUROC & mAUROC & P-mAUROC & mAUROC & P-mAUROC \\
%     % \hline
%     \midrule
%     Hungary Method & -     & 83.85 & 89.31 & 93.32 & 96.19 & 93.82 & 94.67 & 71.51 & 80.56 \\
%     Continuous & -     & 83.40 & 90.51 & 92.85 & 95.91 & 94.33 & 96.03 & 75.01 & 83.71 \\
%     discrete & -     & 85.72 & 93.36 & 94.81 & 96.41 & 94.72 & 96.32 & 75.73 &	85.45\\
%     discrete & direct copy & 85.76 & 93.33 & 94.83 & 96.44 & 94.69 & 96.30 & 75.79	& 85.51 \\
%     discrete & data augment. & \bf 89.21 & \bf 94.07 & \bf 96.75 & \bf 97.45 & \bf 96.71 & \bf 96.99 & \bf 84.45	& \bf 90.45\\
%     \bottomrule[1.2pt]
%     \end{tabular}%
%     }
%   \label{tab:ablation}%
% \end{table*}%

\subsection{Ablation Study}
\noindent\textbf{Component Effectiveness:} We analyze the effectiveness of proposed methods by investigating distribution alignment method, assignment method and target data augmentation. The ablation study carried out on MVTec dataset is presented in Tab.~\ref{tab:ablation}. We make the following observations from the results.
\textbf{i)} KL-Div~\cite{su2023revisiting} and Moment Matching~\cite{liu2021ttt++}-based alignment exhibit the poorest performance across all corruption types. For example, under Gaussian Noise, KL-div achieves only 58.99\% in image-level AUROC. 
This indicates that these alignment methods are not well-suited for handling complex distribution shifts in anomaly detection tasks. In contrast, optimal Transport-based alignment consistently outperforms KL-Div and Moment Matching across all corruptions. 
\textbf{ii)} We further compared with another way of discrete optimal transport solution, i.e. using Hungarian Method to find linear assignment between memory bank and target samples. Introducing Hungarian Method assignment for Optimal Transport yields better results compared to KL-Div. and Moment Matching distribution alignment, as seen in the case of Brighness and Contrast.
% \textbf{ii)} We further compared with another way of discrete optimal transport solution, i.e. using Hungarian Method to find linear assignment between memory bank and target samples. Introducing Hungary Method assignment for Optimal Transport yields better results compared to no assignment strategy, as seen in the case of Gaussian Noise. Similar improvements are observed across other corruptions like Defocus Blur (94.81\% vs. 93.31\%) and Contrast (75.73\% vs. 71.51\%).
\textbf{iii)} {We compare with optimizing the earth moving distance as training objective, denoted as ``Continuous'' under Assignment. The earth moving distance can be derived from the continuous solution, $\gamma$, to optimal transport solution as $\sum_i\sum_j\gamma_{ij}^*C_{ij}$. We observe that using the robust sinkhorn distance~(discretized) is consistently better than the continuous version.}
\textbf{iv)} When directly copying the target data for augmentation, the performance improves further, particularly in pixel-wise AUROC. 
Applying data augmentation instead of direct copying results in the best performance overall. 

In summary, the best-performing configuration combines Optimal Transport alignment with discrete assignment and data augmentation, achieving top scores in both instance-level and pixel-wise AUROC across all corruption types. Notably, the Contrast corruption is still posing great challenge to the method which is explained by the low visibility of defects. %Notably, under Defocus Blur, this configuration achieves a near-perfect pixel-wise AUROC of 97.45\%, and even under more challenging corruptions like Contrast, it maintains high scores (84.45\% pixel AUROC).
In contrast, the KL-Div and Moment Matching methods generally underperform, indicating that more sophisticated distribution alignment techniques, like Optimal Transport, are critical for handling complex distribution shifts in anomaly detection tasks.

% \subsection{Further Analysis}

% \begin{figure}[htbp]
%     \centering
%     \begin{minipage}{0.40\textwidth} 
%         \centering
%         \includegraphics[width=\textwidth]{ICLR 2025/figure/error_rate_comparison.pdf}
%         \vspace{-0.5cm}
%         \caption{Comparison of anomaly sample assignments across different augmentation strategies. }
%         \label{fig:anomaly_assign}
%     \end{minipage}%
%     \hfill
%     \begin{minipage}{0.55\textwidth} 
%         We further demonstrate the effectiveness of the proposed augmentation method by calculating the anomaly samples assigned in the optimal transport solutions. As shown in Figure \ref{fig:anomaly_assign}, out of a total of 2,090 samples in the target domain memory bank, the number of assignments to anomaly samples significantly decreased from 123 (5.89\%) to 79 (3.78\%) when using our data augmentation strategy. This reduction indicates the capability of our method to limit erroneous anomaly sample assignments, thereby improving the optimal transport solution.\\
%         We also evaluated the results of directly copying the target data for augmentation, which slightly reduced the number of assignments to 114 (5.45\%). We believe this minor reduction stems from the larger number of selection pool, but this approach does not smooth the distribution. 
%     \end{minipage}
% \end{figure}

\begin{figure}
% \vspace{-0.5cm}
\centering
        \includegraphics[width=\linewidth]{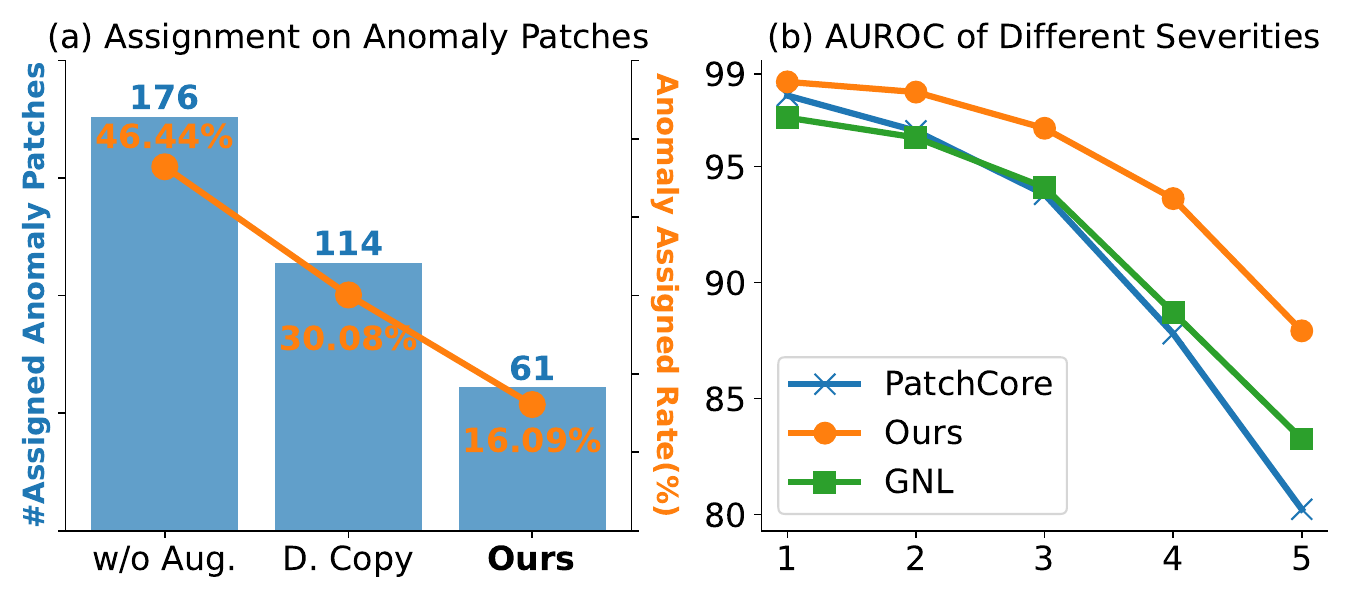}
\vspace{-0.8cm}
    \caption{(a) Comparison of anomaly sample assignments across different strategies. (b) Evaluation on different severity levels of Gaussian Noise Corruption.\vspace{-0.5cm}}
    \label{fig:tardataaug}
\end{figure}

\noindent\textbf{Target Data Augmentation}:
We further demonstrate the effectiveness of the proposed augmentation method by analyzing the anomaly sample assignments in the optimal transport solutions within the same batch. As illustrated in Fig.~\ref{fig:tardataaug} (a), out of total 376 anomaly patches, the number of assigned patches significantly decreased from 176 (46.44\%) to 61 (16.09\%) when applying our data augmentation strategy. This reduction highlights the method’s ability to limit erroneous anomaly assignments, thereby enhancing the quality of the optimal transport solution.
Additionally, we evaluate the impact of simply duplicating the target data for augmentation, which led to a slight reduction in anomaly assignments to 114 (30.08\%). We attribute this minor improvement to the larger selection pool, though this approach fails to smooth the distribution effectively.

% out of 2,090 samples in the source domain memory bank, the number of assignments to anomaly samples in testing data significantly decreased from 123 (5.89\%) to 79 (3.78\%) when applying our data augmentation strategy. This reduction highlights the method’s ability to limit erroneous anomaly assignments, thereby enhancing the quality of the optimal transport solution.
% Additionally, we evaluate the impact of simply duplicating the target data for augmentation, which led to a slight reduction in anomaly assignments to 114 (5.45\%). We attribute this minor improvement to the larger selection pool, though this approach fails to smooth the distribution effectively.

% \begin{table}[htbp]
%   \centering
%   \caption{Numbers of anomaly samples being assigned to target distribution samples}
%     \begin{tabular}{ll}
%           & Anomalous Sample Assigned \\
%     w/o augmentaion & 123/2090 \\
%     w/ augmentation & 72/2090 \\
%     \end{tabular}%
%   \label{tab:robust}%
% \end{table}%

\noindent\textbf{Robustness on Different Severity:} We evaluated the robustness of our method across different corruption severity levels. Fig.~\ref{fig:tardataaug} (b) compares the image level AUROC performance of PatchCore, GNL, and {RoDA} under Gaussian Noise corruption of increasing severity (levels 1 to 5). While all methods exhibit a performance decline as noise severity grows, our method consistently outperforms both PatchCore and GNL at every severity level. Notably, the performance gap becomes more pronounced under higher noise levels, indicating that our approach is more robust to extreme distribution shifts.

% \begin{figure}
% \vspace{-0.5cm}
% \centering
%         \includegraphics[width=\linewidth]{ICCV2025/figure/line_chart.pdf}
% \vspace{-0.3cm}
%     \caption{Evaluation on different severity of Gaussian Noise Corruption. \vspace{-0.5cm}}
%     \label{fig:severity}
% \end{figure}

% \noindent\textbf{Amount of Target Training Data}: We analyze the effect of varying the amount of target training data. As shown in Tab.~\ref{tab:ablation_datasplit}, our paper's default setting uses 20\% of the target domain data for adaptation while the remaining 80\% is reserved for testing. The '80\%' setting represents the reverse split, and the '100\%' setting indicates that all target domain data is used for adaptation, with testing performed on the same set to establish the upperbound of performance. Notably, using only 20\% of the data for adaptation achieves over 98\% of the performance upperbound, demonstrating that our method remains robust even when only a limited amount of target data is available.

\noindent\textbf{Amount of Target Training Data}: We examine the impact of varying the target training data size. As shown in Tab.~\ref{tab:ablation_datasplit}, our default setting uses 20\% of the total target domain data for adaptation, with the remaining 80\% reserved for testing. The ``80\%'' setting reverses this split, while the ``100\%'' setting adapts on the entire target domain data, establishing an upper bound on performance. Remarkably, using just 20\% of the data achieves over 98\% of the upper bound performance, highlighting our method’s robustness even with limited target data.

% Table generated by Excel2LaTeX from sheet 'Sheet1'
\begin{table}[htbp]
  \centering
  \caption{Anomaly detection results on MVTec dataset with different amount of target training data. mAUROC (\%) averaged across all classes.}
  \vspace{-0.2cm}
    \resizebox{0.9\linewidth}{!}{
    \begin{tabular}{l|c|c|c|c|c}
    \toprule[1.2pt]
          $|\set{D}_{tr}|/|\set{D}_t|$& \multicolumn{1}{l|}{G. Noise} & \multicolumn{1}{l|}{D. Blur} & \multicolumn{1}{l|}{Bright.} & \multicolumn{1}{l|}{Contrast} & \multicolumn{1}{l}{Mean}\\
          \midrule
    20\%  & 87.92 & 96.56 & 96.44 & 84.45 & 91.34 \\
    80\%  & 89.40  & 97.82 & 97.70  & 85.01 & 92.48 \\
    100\% & 90.33 & 97.81 & 97.88 & 85.45 & 92.87\\
    \bottomrule[1.2pt]
    \end{tabular}%
    }
  \label{tab:ablation_datasplit}%
  \vspace{-0.4cm}
\end{table}%

%% file: sec/5_conclusion.tex
\section{Conclusion}

In this work, we addressed a realistic challenge of deploying anomaly detection model to target domain with distribution shift. Existing works require modifying training objective and require access source domain data during inference to improve robustness. We relaxed these assumptions by proposing a robust distribution alignment method to mitigate the distribution shift. In particular, a robust Sinkhorn distance is adapted from an existing optimal transport problem to improve the resilience to anomalous patches in the target domain data. We also introduce target data augmentation to reduce the assignment of anomalous patches. We demonstrated the effectiveness on three industrial anomaly detection datasets. The findings suggest future research should pay more attention to the robustness of anomaly detection under realistic challenges.